\pdfoutput=1

\documentclass[conference]{IEEEtran}
\usepackage{times}

\usepackage[numbers]{natbib}
\usepackage{multicol}
\usepackage[bookmarks=true]{hyperref}

\usepackage{amsmath,amsfonts}
\usepackage{algorithmic}
\usepackage{algorithm}
\usepackage{graphicx}
\usepackage{amssymb}
\usepackage{makecell}

\usepackage{xcolor}

\usepackage{flushend}

\usepackage{epstopdf}

\makeatletter

\newcommand{\Rmnum}[1]{\expandafter\@slowromancap\romannumeral #1@}
\makeatother

\begin{document}

\title{Learning-Based Adaptive Control for Surgical Robotic Exposure Task on Deformable Tissues}

\author{
\authorblockN{
Jiayi Liu,
Kaiqi Wei,
Yiwei Wang\authorrefmark{1}, 
Huan Zhao and
Han Ding
}
\authorblockA{Huazhong University of Science and Technology, Wuhan, China}
\authorblockA{\authorrefmark{1}Corresponding to: {\tt\small wang\_yiwei@hust.edu.cn}}}

\maketitle

\begin{abstract}
In various surgical procedures, regions of interest (ROIs) such as organs or lesions are often occluded by overlying tissues, requiring surgeons to achieve adequate exposure for precise intervention. However, the irregular geometry, nonlinear biomechanical properties of overlying tissues, and limited intraoperative visibility of the ROI pose significant challenges to the autonomous execution of tissue retraction. To address this, we formulate a realistic model of the tissue retraction task and propose a learning-based adaptive control framework for achieving ROI exposure. The method optimizes control inputs online by monitoring changes in the visual boundary of the tissue, while leveraging a deep deformation estimation model trained on simulation data to identify the optimal grasping point and ensure the convergence and safety of the adaptive controller. Through simulations and real-world experiments on different deformable materials, it has been demonstrated that this framework exhibits zero-shot adaptation to similar tasks and can complete the autonomous retraction process, from initial grasp selection to full ROI exposure. Therefore, it has the potential to be applied in actual surgical assistance scenarios. 
\end{abstract}

\IEEEpeerreviewmaketitle

\section{Introduction}

Robotic-assisted surgery (RAS) has achieved remarkable progress in clinical practice in recent years, significantly enhancing surgical outcomes across diverse procedures while effectively reducing the workload of medical personnel \cite{chatterjee2024advancements}. As research continues to advance the autonomy of surgical systems, an increasing number of auxiliary tasks demonstrate potential for independent execution by robotic platforms, thereby further liberating human surgical assistants \cite{saeidi2022autonomous}. Among these, tissue retraction is the most common auxiliary procedure in various surgeries \cite{patil2010toward}. For example, during a laparoscopic cholecystectomy, the gallbladder must be retracted to adequately expose the Calot’s triangle. Currently, such retraction remains heavily dependent on close coordination between surgical assistants and the primary surgeon. The autonomy of this operation will significantly reduce the physical and mental exertion of medical staff \cite{attanasio2020autonomous}, but it faces some challenges: the irregular shape of occluding tissues impedes precise modeling, the highly nonlinear deformation behavior of biological tissues resists accurate prediction, and the region of interest (ROI) cannot be precisely determined before full exposure. These fundamental issues make autonomous tissue retraction one of the key focuses of current surgical robotics research.

To address these challenges, recent research in surgical robotics has increasingly focused on integrating advances in perception, modeling, and control to enable autonomous \cite{long2025surgical} or semi-autonomous \cite{attanasio2020autonomous, schussler2025semi} tissue retraction. Early approaches \cite{patil2010toward} relied on prior physics-based simulation models for optimization-based planning, which limited generalizability due to their strong dependence on these prior models. Subsequently, learning-based methods \cite{chen2023novel, chen2025autonomous, pore2021safe, pore2021learning, scheikl2022sim} that combine imitation learning and reinforcement learning have achieved end-to-end control, improving sample efficiency and safety through mechanisms such as contrastive representation and asymmetric input. Meanwhile, to enhance clinical interpretability, logic-based frameworks like FRAS \cite{meli2021autonomous} and DEFRAS \cite{tagliabue2022deliberation} explicitly encoded surgical domain knowledge using answer set programming, enabling failure-aware replanning and robust handling through online model updating. Moreover, DeformerNet \cite{thach2022learning, thach2023deformernet, thach2025deffusionnet} has been proposed as a point cloud–driven shape servoing framework that outputs end-effector displacement commands by learning the low-dimensional embedding difference between the current and target shapes, thereby enabling tissue retraction without requiring any physical prior model. Similarly, shape servoing forms the core technical foundation of this paper, as a general deformation control method.

\begin{figure}[!t]
\centering
\vspace{+0.15cm}
\includegraphics[width=3.49in]{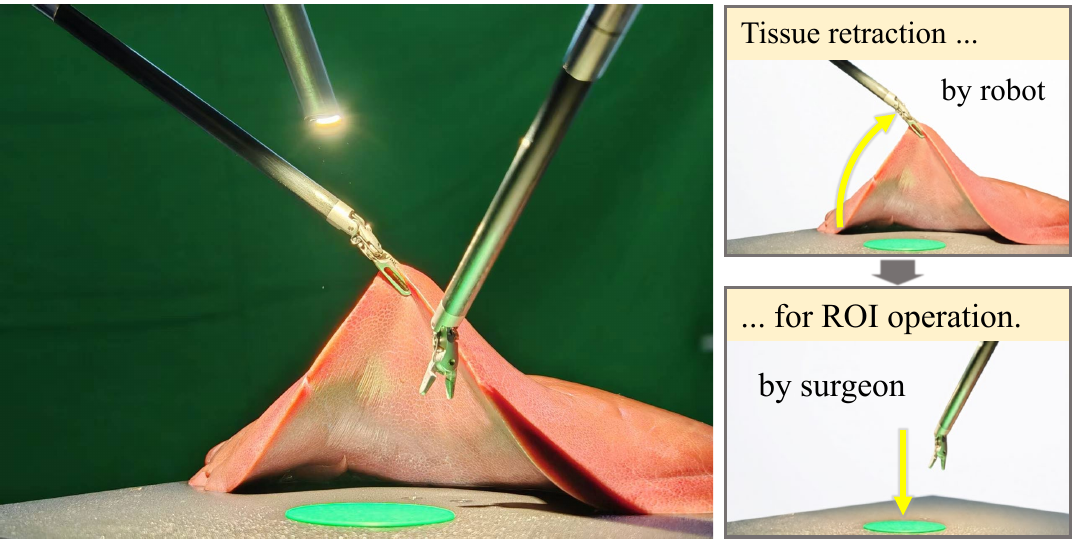}
\vspace{-0.7cm}
\caption{The robot performs tissue retraction to expose the ROI, thereby enabling the surgeon to perform intervention on the obstructed area.}
\vspace{-0.6cm}
\label{fig_1}
\end{figure}

Our method for shape servoing of deformable tissues revolves around the deformation Jacobian. This idea was first proposed by Navarro-Alarcón et al. \cite{navarro2013model, navarro2016automatic}, who proposed a model-free adaptive shape control framework by estimating and updating the deformation Jacobian matrix of unknown elastic objects online. To enhance robustness, Lagneau et al. \cite{lagneau2020active, lagneau2020automatic} employed a sliding-window weighted least-squares method combined with an eigenvalue-based confidence criterion to update the Jacobian, effectively mitigating the impact of observation noise. Liu et al. \cite{liu2025autonomous} introduced a deep reinforcement learning agent to guide the adjustment of the Jacobian, improving convergence performance. Separately, to address the problem of target point displacement caused by tissue deformation in surgical scenarios, Alambeigi et al. \cite{alambeigi2018toward} and Zhong et al. \cite{zhong2019dual} proposed estimating the deformation mapping under probe-tissue interaction to move the target point to a fixed instrument position in renal tumor cryoablation and suturing tasks, respectively. Recently, research has extended toward more sophisticated deformation representations. Yang et al. \cite{yang2023model, yang2023modal} constructed low-dimensional, physically interpretable global deformation features based on modal analysis and derived an analytical form of the deformation Jacobian. Yu et al. \cite{yu2022global} integrated offline pre-training with online adaptation, leveraging neural networks to learn a global Jacobian model for large-deformation control of deformable linear objects.  Collectively, these advances demonstrate that adaptive control methods based on online deformation Jacobian estimation remain an effective paradigm for shape servoing tasks.

In this paper, we propose a unified framework that integrates learning and adaptive control for autonomous robotic tissue retraction tasks, with the explicit goal of exposing the ROI, as illustrated in Fig. \ref{fig_1}. First, we design a set of low-dimensional features that effectively represent occlusion boundaries, and introduce the deformation Jacobian in this feature space to construct a model-free adaptive deformation control method. Then, we build a simulation for this retraction task to explore the interaction and deformation relationship between the manipulator and the tissue, and use this to train a data-driven deformation estimation model. The introduction of this model aims to improve the stability, reliability, and safety of the controller in actual operation. Importantly, we focus on the zero-shot transfer capability of the proposed method in real-world tasks, meaning that prior knowledge from simulation can be effectively integrated into the online adaptive adjustment process without retraining or fine-tuning in the real environment, thus achieving efficient and reliable retraction of unknown tissues.

The main contributions of our work are as follows:
\begin{itemize}
    \item A novel tissue retraction framework is proposed. It estimates the deformation Jacobian by observing the tissue's boundary, and then uses it to drive the control system.
    \item A data-driven deep deformation estimation model is trained on simulation and is integrated to provide prior knowledge for the adaptive control pipeline.
    \item Real-world experiments are designed and conducted on ex vivo tissues to validate the effectiveness and practical application potential of our proposed method.
\end{itemize}

\section{Task Formulation}

\begin{figure}[!t]
\centering
\vspace{+0.2cm}
\includegraphics[width=3.49in]{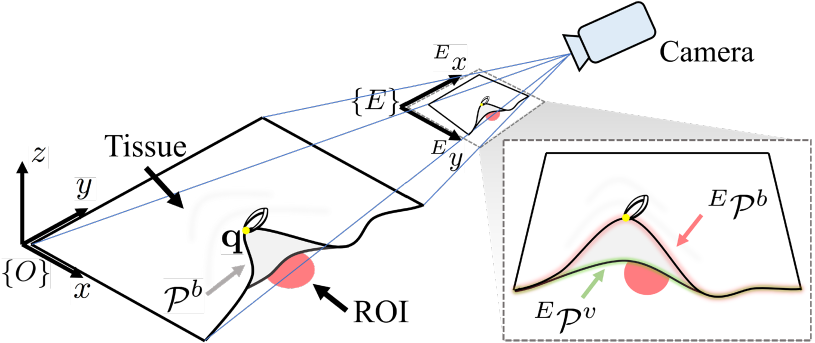}
\vspace{-0.4cm}
\caption{Task formulation and key definitions for tissue retraction.}
\vspace{-0.4cm}
\label{task}
\end{figure}

To model the tissue retraction task aimed at exposing an ROI while closely approximating the scenarios encountered in real surgical retraction procedures, this paper proposes the task formulation shown in Fig. \ref{task}. Specifically, a deformable elastic sheet is used to broadly represent the occluding tissue, with an irregularly shaped boundary to reflect the morphological diversity of real biological tissues. Additionally, a customizable two-dimensional elliptical marker with adjustable shape, scale, position, and orientation is introduced to simulate the ROI that needs to be exposed in different surgical scenarios.

The retraction action $\mathbf{q}$ can be applied to any point along the tissue boundary $\mathcal{P}^b$ to produce deformation. Sensory feedback is derived from the 2D image provided by a laparoscopic camera that can be positioned at any location and orientation to capture the tissue. Here, ${}^E\mathcal{P}^b$ represents the projection of $\mathcal{P}^b$ onto the image plane, while ${}^E\mathcal{P}^v$ represents the visual boundary under the current actual occlusion state. The task is considered completed when ${}^E\mathcal{P}^v$ no longer overlaps with the projection of the ROI on the image plane, meaning the ROI is fully exposed and unobstructed in the field of view.

\section{Methodology}

\begin{figure}[!t]
\centering
\vspace{+0.2cm}
\includegraphics[width=3.49in]{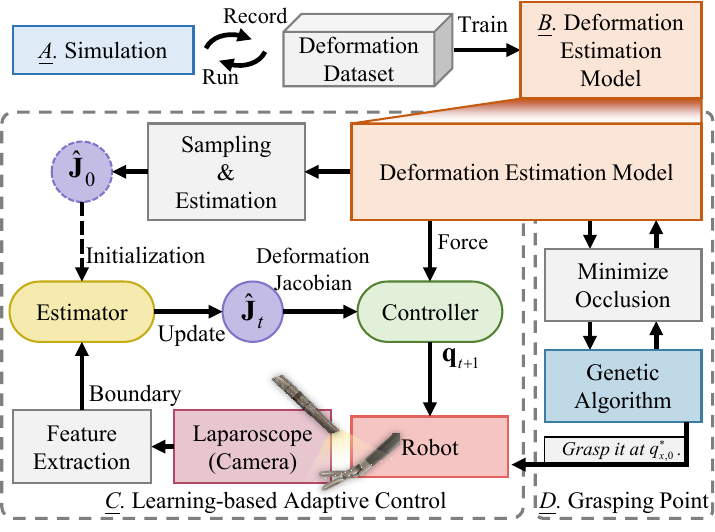}
\vspace{-0.6cm}
\caption{Tissue retraction pipeline: The deformation estimation model is trained using data collected from simulations. Before the robotic retraction, it is employed to optimize the grasping point and initialize the deformation estimator. During retraction, it estimates deformation forces to serve as safety constraints for the controller. Tissue retraction is accomplished through continuous adjustment and execution within a closed-loop adaptive control.
}
\vspace{-0.4cm}
\label{framework}
\end{figure}

This paper proposes a learning-based adaptive control method to accomplish the task of retracting occluding tissues for ROI exposure. Specifically, we first train a deformation estimation model in simulation. This model serves to initialize the deformation Jacobian and provide virtual deformation force constraints during the control iterations. Furthermore, we discuss the selection of the grasping position. Fig. \ref{framework} shows the overall framework of the proposed method.

\subsection{Physics-Based Deformable Object Simulation}

In order to collect prior deformation data, a simulation environment corresponding to the task formulation needs to be constructed.
To balance accuracy and efficiency, the mass-spring model (MSM) \cite{mollemans2003tetrahedral, qiao2009research} is adopted for deformation computation. In this model, mesh vertices are treated as point masses, and edges between adjacent vertices are modeled as springs. The elastic force between two connected masses $i$ and $j$ follows Hooke’s law:
\begin{equation}
    \mathbf{F}_{ij} = k(\| \mathbf{p}_i - \mathbf{p}_j \| - l_0) \mathbf{e}_{ij},
    \label{Hook}
\end{equation}
where $k$ is the stiffness, $l_0$ is the rest length, and $\mathbf{e}_{ij}$ is the unit direction vector. The system is integrated using explicit Euler with timestep $ \Delta t$ and velocity damping $\zeta \in (0,1]$:
\begin{equation}
    \dot{\mathbf{p}} \leftarrow (\dot{\mathbf{p}} + \mathbf{F}/m_{eq} \cdot \Delta t) \cdot \zeta, \quad \mathbf{p} \leftarrow \mathbf{p} + \dot{\mathbf{p}} \cdot \Delta t ,
\end{equation}
where $\mathbf{p}$ and $\dot{\mathbf{p}}$ denote the positions and velocities, respectively, of the non-fixed masses. The deformable object’s shape is updated iteratively using this integration rule, which is achieved through the hardware acceleration of Taichi \cite{hu2019taichi}.

Since the selection of the actual retraction strategies is influenced by the tissue boundary shape, the soft tissue model constructed for this simulation has randomized and irregular boundaries. The finite sine series $f(x) = \sum_{k=1}^{K} a_k \cdot \sin(k \pi x), \quad x \in [0, 1]$ is used to generate randomized boundaries, where $K \in \mathbb{Z}_{>0}$ is the number of harmonic terms controlling shape complexity, and $a_k \sim \mathcal{U} (-0.1,0.1)$ are random amplitudes. This boundary profile function is scaled and applied to carve into a base cuboid grid of dimensions $100\times100\times4$, generating an occluding tissue model with a smoothly curved boundary.

In the task, the controlled variable is the position and orientation of the instrument, which is represented as $\mathbf{q} = [q_x, q_y, q_z, q_\alpha, q_\beta, q_\gamma]^T\in\mathbb{R}^6$. The initial grasping position $\mathbf{q}^p_0 = [q_{x,0}, q_{y,0}, q_{z,0}]^T \in \mathcal{P}^b_0$ is chosen arbitrarily along the boundary, where $\mathcal{P}^b_0$ denotes the set of positions on the initial tissue boundary. The instrument begins at the initial pose $ \mathbf{q}_0 = [(\mathbf{q}^p_0)^T, (\mathbf{0})^T] ^T$. Grid nodes in the vicinity of the instrument’s grasping point are rigidly displaced according to its motion, while the positions of all other non-fixed nodes are updated iteratively through the MSM.

\subsection{Deep Deformation Estimation Model}

During the preparation phase, we aim to train a deep neural network endowed with fundamental deformation knowledge, enabling it to guide the control process across a variety of similar retraction scenarios and compensate for deformation force information that is not readily measurable in the real world. The simulation environment is used to generate a dataset of action–deformation correspondences through random interaction, and the intermediate process of the MSM algorithm provides the relative values of the force vectors.

To represent the boundary $\mathcal{P}^b$, we employ an expansion in terms of Wendland radial basis functions \cite{wendland1998error}:
\begin{equation}
    \phi(\xi) = \sum_{j=1}^{m} w_j \, \varphi_{3,1} \left( \left|\frac{\xi - \xi_c^j}{h}\right| \right),
    \label{eq:rbf_boundary}
\end{equation}
where $\xi_c^j$ denotes the center of the $j$-th kernel, $h>0$ is the support radius, and $\varphi_{3,1}$ is the Wendland kernel yielding a $C^2$-continuous, compactly supported basis function. The number of kernels is denoted by $m \in \mathbb{Z}_{>0}$. Owing to their locality, smoothness, and numerical stability, Wendland kernels have been widely used in interpolation \cite{de2007mesh}, function approximation \cite{kouibia2024approximation}, and soft tissue deformation modeling \cite{wachowiak2004compact}.
Importantly, both the derivatives and integrals of $\varphi_{3,1}$ admit closed-form expressions, which greatly simplify subsequent analytical computations. Using this kernel, we parameterize the 3D tissue boundary via a dedicated operator:
\begin{equation}
    \mathbf{W}^b = \mathcal{R}^b (\mathcal{P}^b),
\end{equation}
where $\mathcal{R}^b(\cdot)$ denotes the parameterization mapping that converts the tissue boundary $\mathcal{P}^b$ into a compact coefficient matrix $\mathbf{W}^b \in \mathbb{R}^{m \times3}$. Full details of this construction, including kernel configuration, data centering, and least-squares fitting, are provided in Appendix \Rmnum{1}.

The maximum deformation force is of crucial importance for designing safety constraints. It is assumed to occur at grasped nodes $ \mathcal{P}^g$. Since the MSM computes pairwise forces between mesh nodes during deformation simulation according to Equation (\ref{Hook}), the resultant force on each grasped node can be calculated via summation over neighboring nodes, and the maximum deformation force can be extracted by:
\begin{equation}
    F^{\max} = \max_{\mathbf{p}^i \in \mathcal{P}^g} \left\| \sum\nolimits_{\mathbf{p}^j \in \mathcal{N}(i)} (\mathbf{F}_{ij}) \right\|_2,
\end{equation}
where $\mathcal{N}(i)$ denotes the set of neighboring nodes of the $i$th node. The prediction model, while predicting deformation, also estimates the resulting deformation forces serving as a perception complement in scenarios without force sensors.

To achieve real-time prediction and estimation of shape and force, a multi-layer perceptron (MLP) with dimensions $(m{+}7){-}64{-}256{-}512{-}256{-}(3m{+}1)$ is employed to construct the deep deformation estimation model:
\begin{equation}
[\mathrm{vec}(\mathbf{W}^b)^T, F^{\max}]^T = \text{MLP}_\theta([\mathbf{w}_{x,0}^T, \mathbf{q}^T, q_{x,0}]^T),
\end{equation}
where $\mathrm{vec}(\mathbf{W}^b)$ represents the vectorization of $\mathbf{W}^b$ with $3m$ parameters. $\mathbf{w}_{x,0} \in \mathbb{R}^m$ denotes the column vector of $\mathbf{W}^b_0 = [\mathbf{w}_{x,0}, \mathbf{0}, \mathbf{0}]$ corresponding to the x-coordinate component. Since the initial boundary point set $\mathcal{P}^b_0$ lies on the x-y plane, the boundary shape can be parameterized by x along the y-direction, reducing the number of required parameters. After reconstructing the boundary curve, the grasping position constrained to lie on the curve can be uniquely determined by specifying $q_{x,0}$ alone. 

\subsection{Learning-Based Adaptive Control}

The learning-based adaptive control method leverages the trained deformation model to guide actions and enforce safety constraints. The retraction task in this phase is defined in a 2D coordinate frame $\{E\}$, which lies in the plane of the camera image. The x-axis of $\{E\}$ is aligned with the line connecting the start and end points of the observed tissue boundary ${}^E\mathcal{P}^b$, while the y-axis is oriented parallel to the camera’s vertical axis. The origin of $\{E\}$ is placed at the starting point of the observed tissue boundary. ${}^E\mathcal{P}^b$ is defined as the projection of the tissue boundary onto the image plane. In the early stage of retraction, it is almost consistent with ${}^E\mathcal{P}^v$.

To achieve the exposure task in closed-loop control using only camera image feedback, we focus on the visual boundary ${}^E\mathcal{P}^v$. Due to differences in dimensionality and in the sampling strategy used to extract coordinates from the simulation, we employ a separate parameterization operator:
\begin{equation}
    \mathbf{w}^v = \mathcal{R}^v ({}^E\mathcal{P}^v),
\end{equation}
where $\mathcal{R}^v (\cdot)$ denotes the parameterization with full details shown in Appendix \Rmnum{2}. The resulting parameter vector $\mathbf{w}^v \in \mathbb{R}^m$ encodes the shape of ${}^E\mathcal{P}^v$. 

The following part presents the modeling of the retraction task. It is essential to clarify that this task is exploratory. Before the ROI is exposed, retraction decisions can only be made based on the currently visible regions. The primary objective is to reduce the occlusion area of the overlapping parts between the tissue and the ROI. The loss function reflecting the occlusion is designed as follows:
\begin{align}
    \mathcal{L} &= \int_\mathcal{D} \phi^v({}^Ex) d{}^Ex \nonumber \\
    &= \int_\mathcal{D} \sum^m_{i=1}{}^i\mathbf{w}^v \varphi \left( \left | \frac{{}^Ex - {}^Ex_c^i}{h} \right | \right) d{}^Ex,
    \label{L}
\end{align}
where $\mathcal{D} = \cup_s[x_{Ls}, x_{Hs}]$ represents the overlap area between the boundary of the observable ROI and that of the upper tissue. The overlap area is generally a single segment, that is, $s = 1$. There could also be multiple segments, that is, $s > 1$. Furthermore, if the ROI is fully covered, the integration area is set to the entire domain.

The control objective is to minimize the loss function in the direction of the optimal action, achieving efficient exposure. The partial derivative of $\mathcal{L}$ with respect to $\mathbf{q}$ is computed to obtain the pose update for the instrument:
\begin{equation}
    \frac{d}{dt} \mathbf{q} = -\kappa \frac{\partial\mathcal{L}}{\partial\mathbf{q}} = - \kappa \left(\frac{\partial\mathbf{w}^v}{\partial\mathbf{q}}\right)^T \frac{\partial\mathcal{L}}{\partial\mathbf{w}^v},
    \label{dq/dt}
\end{equation}
where $\kappa$ is a positive constant value. The second term captures the relationship between the occluded area and the visual boundary parameters, and can be computed by combining it with Equation (\ref{L}):
\begin{equation}
    \frac{\partial \mathcal{L}}{\partial \mathbf{w}^{v}} =
    \begin{bmatrix}
    \displaystyle \int_\mathcal{D} \varphi \left( \left| \frac{{}^Ex - {}^Ex_c^i}{h} \right| \right) d{}^Ex
    \end{bmatrix}_{i=1}^m.
    \label{L/w}
\end{equation}
An adaptive deformation Jacobian $\mathbf{J} \in \mathbb{R}^{m \times 6}$ is introduced to describe the relationship between visual boundary parameters and the instrument's actions. Based on $\mathbf{J}$, the first term in Equation (\ref{dq/dt}) can be determined as: 
\begin{equation}
    \frac{\partial\mathbf{w}^v}{\partial\mathbf{q}} = \mathbf{J}.
\end{equation}

To obtain an accurate real-time Jacobian estimate, it must be reasonably initialized during the setup phase to ensure convergence. Subsequently, during control iterations, the Jacobian is adaptively adjusted based on the observed estimation error. The deformation estimation model is used to participate in the estimation of $\mathbf{J}_0$. First, $\nu$ sets of random actions are generated $\mathbf{Q}_0 = [\delta\mathbf{q}^1_0, \delta\mathbf{q}^2_0, \cdots, \delta\mathbf{q}^\nu_0] \in \mathbb{R}^{6 \times \nu}$. The model infers the corresponding boundary parameters for each action based on the grasping position and the initial boundary parameters. The boundary deformation parameters estimated by the model are as follows:  
\begin{align}
    &{}^E\hat{\mathbf{W}}_0^b = [\delta{}^E\mathbf{w}_0^{b,1}, \cdots, \delta{}^E\mathbf{w}_0^{b,\nu}] \in \mathbb{R}^{m \times \nu},
    \\
    &\delta{}^E\mathbf{w}_0^{b,i} = {}^E\mathcal{T}(\hat{\mathbf{W}}^{b,i}_0)-{}^E\mathbf{w}_0^b, \quad i \in \{1, \dots, \nu\},
\end{align}
where $\hat{\mathbf{W}}^{b, i}_0$ represents the boundary parameters being estimated by the model in the $\{O\}$ coordinate, and ${}^E\mathcal{T}(\cdot)$ is the function that transforms it into the $\{E\}$ coordinate. Since around the initial state, the visual boundary ${}^E\mathcal{P}^v$ is close to the tissue boundary ${}^E\mathcal{P}^b$, it can be approximately considered that ${}^E\hat{\mathbf{W}}_0^v \approx {}^E\hat{\mathbf{W}}_0^b$. Then the initial Jacobian can be estimated by the relationship $\langle {}^E\hat{\mathbf{W}}_0^b, \mathbf{Q}_0 \rangle$:
\begin{equation}
    {}^E\hat{\mathbf{W}}_0^b = \hat{\mathbf{J}}_0 \mathbf{Q}_0.
\end{equation}
This can be formulated as a least-squares problem:
\begin{equation}
    \underset{\hat{\mathbf{J}}_0}{\mathrm{minimize}}\| \mathbf{Q}_0^T \hat{\mathbf{J}}_0^T - ({}^E\hat{\mathbf{W}}_0^b)^T \|^2_F.
\end{equation}
Its analytical solution is given by:
\begin{equation}
    \hat{\mathbf{J}}_0 = {}^E\hat{\mathbf{W}}_0^b \mathbf{Q}_0^T \left( \mathbf{Q}_0 \mathbf{Q}_0^T \right)^{-1},
\end{equation}
where $\mathbf{Q}_0 \mathbf{Q}_0^T$ is a nonsingular matrix because it is from random sampling, hence, the quadratic form can be inverted.

During manipulation, the deformation Jacobian changes, so the estimated values need to be adjusted. During each control iteration, the parameter changes resulting from the execution of the action can be estimated as follows:
\begin{equation}
    \delta \hat{\mathbf{w}}_{t-1}^v = \hat{\mathbf{J}}_{t-1} \delta \mathbf{q}_{t-1}.
\end{equation}
The actual visual boundary parameters can be observed through the image. The error between the theoretical estimation and the actual parameters is denoted by $\mathbf{e}$:
\begin{equation}
    \mathbf{e} = \hat{\mathbf{J}}_{t-1} \delta \mathbf{q}_{t-1} - \delta \mathbf{w}_{t-1}^v \in \mathbb{R}^m.
\end{equation}
To maintain the estimated $\hat{\mathbf{J}}$ close to the deformation model, the algorithm adapts to match the estimated error. It employs a gradient descent method to minimize a quadratic function:
\begin{equation}
    \mathcal{H} = \mathbf{e}^T\mathbf{e}/2.
\end{equation}
Update $\hat{\mathbf{J}}$ according to the gradient rule:
\begin{equation}
    \frac{d}{dt} {}^i \hat{\mathbf{J}} = -\lambda \frac{\partial \mathcal{H}}{\partial {}^i \hat{\mathbf{J}}} = - \lambda {}^i \mathbf{e} \delta \mathbf{q}^T_{t-1},
\end{equation}
where $\lambda > 0$ is the adaptation rate. ${}^i \hat{\mathbf{J}}$ represents the row vector of $\hat{\mathbf{J}}$, and ${}^i \mathbf{e}$ denotes the element of $\mathbf{e}$. In the discrete control iterations, the update rule takes the form of:
\begin{equation}
    \hat{\mathbf{J}}_t = \hat{\mathbf{J}}_{t-1} - \lambda \mathbf{e} \delta \mathbf{q}^T_{t-1}.
\end{equation}

In practical retractions, excessive deformation forces may cause tissue damage. Therefore, force constraints are imposed on instrument manipulation. Building upon the loss function in Equation (\ref{L}), a force penalty term is added:
\begin{equation}
    \mathcal{L}^* = \mathcal{L} + \tau \mathcal{L}^F \left( \hat{F}^{\max} - F^{\mathrm{safe}} \right),
\end{equation}
where $\tau$ denotes the weight of the force penalty. $\hat{F}^{\max}$ is the maximum deformation force estimated by the network for $\mathbf{q}_t$. $F^{\mathrm{safe}}$ represents the maximum force threshold that the tissue can safely withstand. The form of $\mathcal{L}^F\left(x\right)$ is as follows:
\begin{equation}
    \mathcal{L}^F\left(x\right) = \left\{ \begin{array}{rcl}
    0\ ,     & \mbox{if} & x\leq0,\\
    x^2,   & \mbox{if} & x>0.
    \end{array} \right.
\end{equation}
It can be seen that when the estimated force is less than the safety value, the additional term is $0$, so the action update is consistent with Equation (\ref{dq/dt}). However, when the estimated force exceeds the safety value, the update rule becomes:
\begin{equation}
    \frac{d}{dt} \mathbf{q} = -\kappa_1 \frac{\partial\mathcal{L}}{\partial\mathbf{q}} - \kappa_2 \frac{\partial\mathcal{L^F}}{\partial\mathbf{q}},
\end{equation}
where $\kappa_2 = \tau\kappa_1$. The method for solving the first partial derivative is the same as that mentioned earlier. The partial derivative of the penalty term is solved as follows:
\begin{align}
    \frac{\partial\mathcal{L}^F}{\partial\mathbf{q}} 
    &= \frac{\partial \mathcal{L}^F}{\partial \hat{F}^{\mathrm{max}}} \cdot \frac{\partial \hat{F}^{\mathrm{max}}}{\partial \mathbf{q}} \nonumber \\
    &= 2 \left( \hat{F}^{\mathrm{max}} - F^{\mathrm{safe}} \right) \cdot \mathrm{MLP}^{-1}_\theta\!\left( \hat{F}^{\mathrm{max}} \right),
\end{align}
where $\text{MLP}^{-1}_\theta(\cdot)$ represents the backpropagation process, which enables the calculation of the gradient relationship between the input and output of the neural network.

To form the controller, $\mathbf{u}_t = d\mathbf{q}/dt = [(\mathbf{u}_t^v)^T, (\mathbf{u}_t^\omega)^T]^T$ is used to represent the desired increment of the instrument's position. The maximum speeds for both position and orientation need to be restricted separately:
\begin{equation}
    \mathbf{u}^{v/\omega,*}_t = \min \left( \mathbf{u}^{v/\omega}_t, s^{v/\omega} \cdot \mathbf{u}^{v/\omega}_t / \| \mathbf{u}^{v/\omega}_t \|_2 \right) ,
\end{equation}
where scalar $s^{v/\omega}$ determines the maximum linear or angular velocity during the robot’s manipulation. The function $\min(\cdot)$ returns the vector with the smaller modulus from the two vectors provided. The final controller is as follows:
\begin{equation}
    \mathbf{q}_{t+1} = \mathbf{q}_t + [(\mathbf{u}^{v,*}_t)^T, (\mathbf{u}^{\omega,*}_t)^T]^T.
\end{equation}
The robot will control the grasper according to this rule until the ROI is fully exposed ($\mathcal{D} = \varnothing$).

\subsection{Preoperative Grasping Point Optimization}

Different grasping positions significantly affect the retraction action and exposure efficiency, and once selected, the grasping position cannot be changed during the continuous retraction process. Therefore, choosing an appropriate grasping position is another key issue that needs to be addressed.

To obtain the optimal grasping position under partial observations, the idea is to find the position where the action performed maximizes the exposure of the initial observed ROI. Additionally, the deformation produced by this action must be within the safe range. An optimization problem with inequality constraints is formulated as follows:
\begin{align}
    & \underset{q_{x,0}, \mathbf{q}}{\mathrm{minimize}} \left(\hat{\mathbf{w}}^v(q_{x,0}, \mathbf{q}) - {}^E\mathbf{w}^b_0 \right)^T \left(\epsilon \frac{\partial \mathcal{L}_\mathcal{D}}{\partial \mathbf{w}^v} + (1-\epsilon) \frac{\partial \mathcal{L}_\Omega}{\partial \mathbf{w}^v} \right), \\
    & \text{s.t.} \quad \hat{F}^{\max}(q_{x,0}, \mathbf{q}) < F^\mathrm{safe} ,
\end{align}
where $\mathcal{L}_\mathcal{D}$ and $\mathcal{L}_\Omega$ respectively represent the shaded area on the boundary of coverage and the shaded area on the entire boundary. The hyperparameter $\epsilon$ is used to balance the two aspects. A larger $\epsilon$ emphasizes the observed ROI exposure, while a smaller value emphasizes the overall exposure. In the constraint, the deformation estimation model takes in the action and provides an estimated deformation force, which is limited to be less than the safe force. This optimization demonstrates the process of selecting the action that achieves the best exposure with a limited retracting force.

\begin{figure}[!t]
\centering
\vspace{+0.1cm}
\includegraphics[width=3.0in]{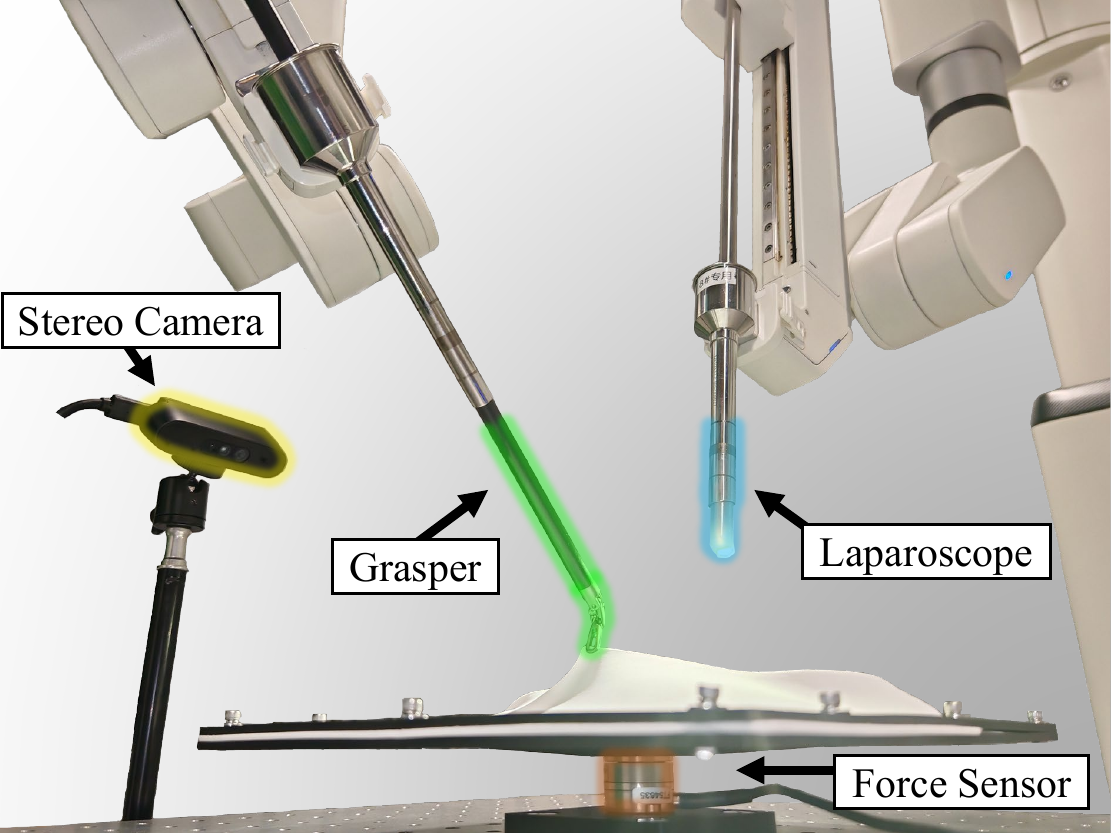}
\vspace{-0.2cm}
\caption{Real-world experimental setup.}
\vspace{-0.5cm}
\label{exp_setup}
\end{figure}

\begin{figure*}[!t]
\centering
\begin{minipage}{0.31\linewidth}
    \centering
    \includegraphics[height=2.8in]{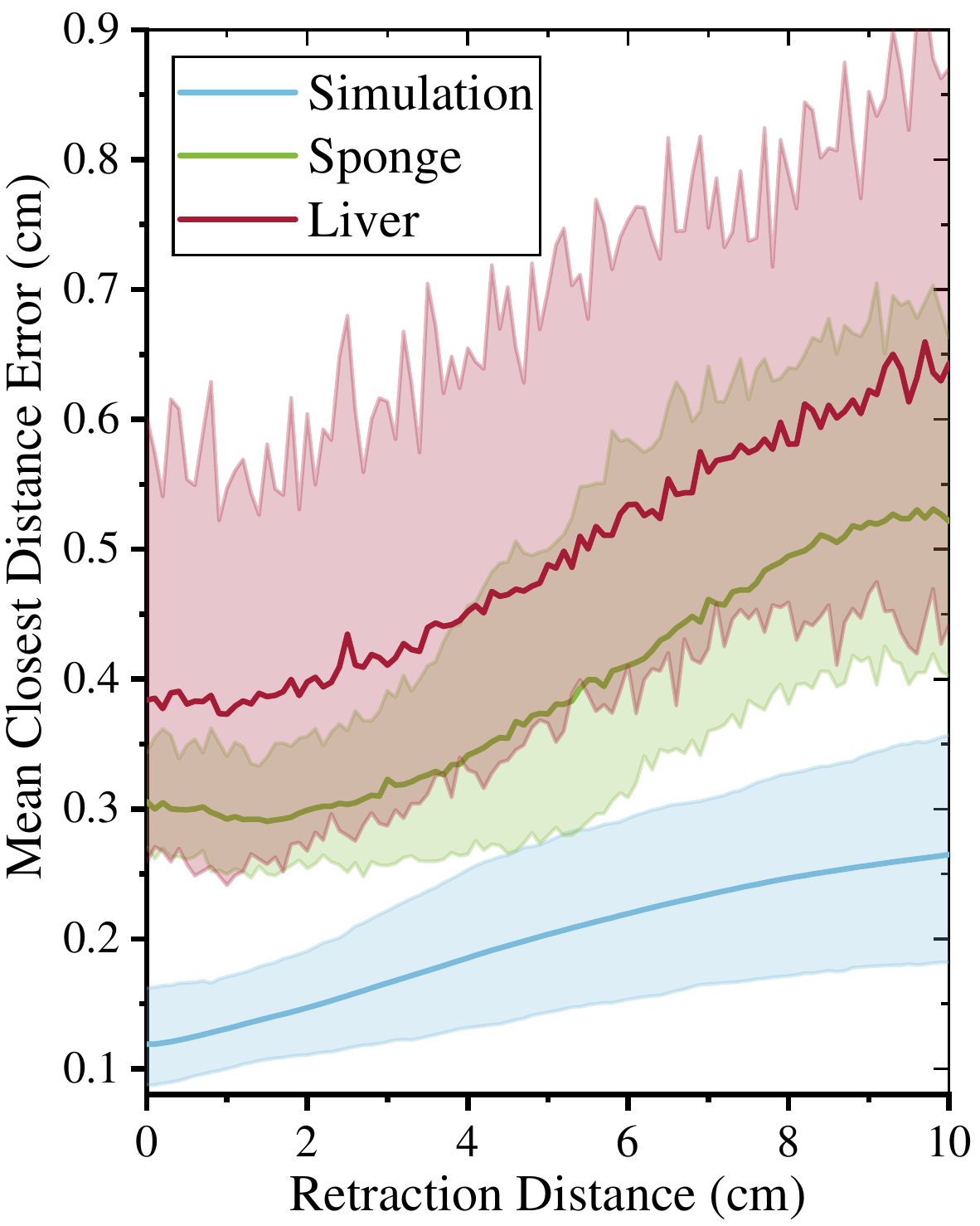}
\end{minipage}
\begin{minipage}{0.682\linewidth}
    \centering
    \vspace{+0.15cm}
    \includegraphics[height=2.81in]{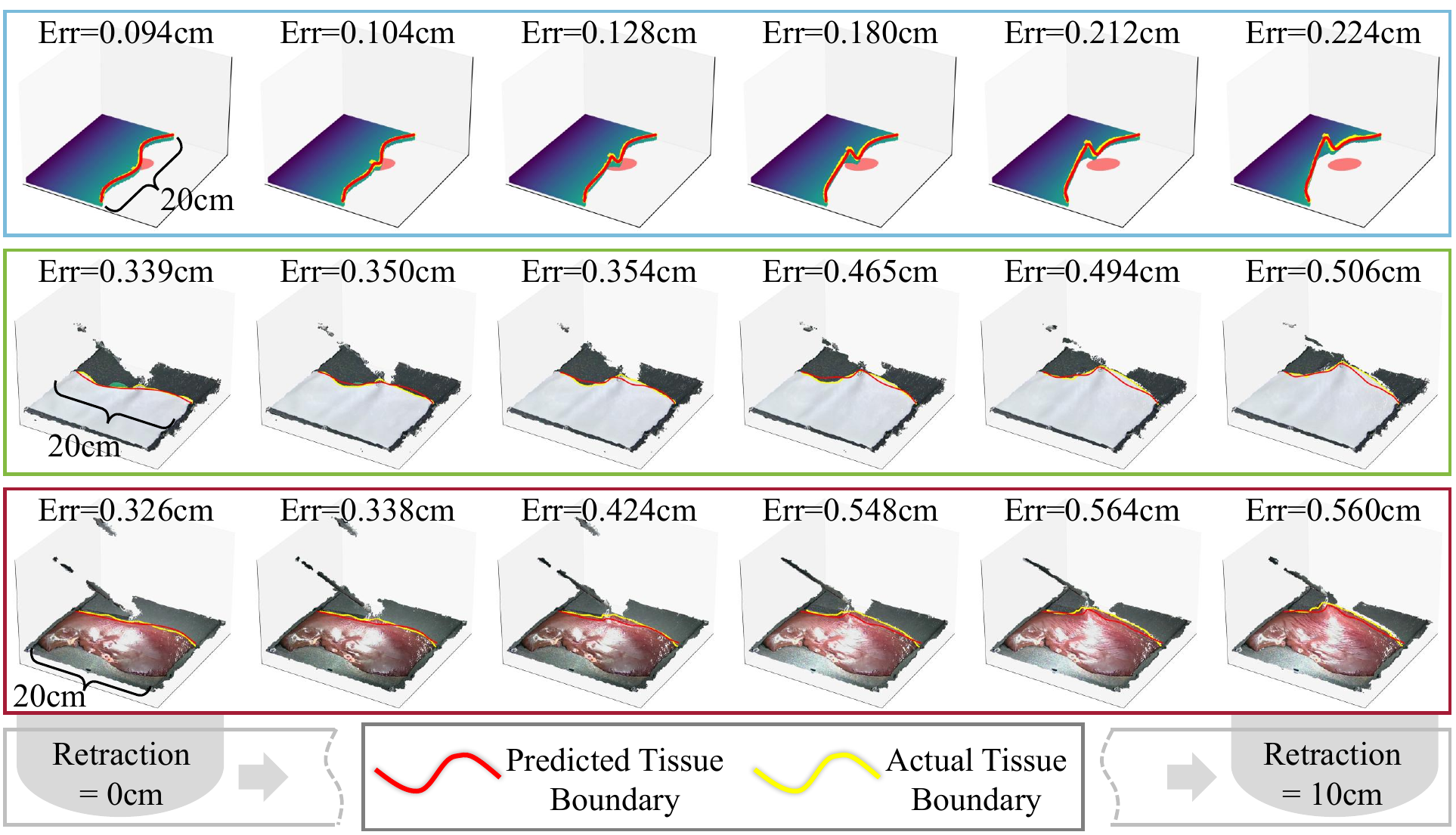}
\end{minipage}
\vspace{-0.2cm}
\caption{(Left) The mean closest distance error, which is computed between the deformation network’s predicted boundaries and the measured boundaries in relation to the retraction distance across the three experimental scenarios. For each retraction distance, the solid line denotes the mean error, and the translucent band represents the 10th to 90th percentile range. (Right) A retraction process is visualized for each experimental scenario, with the predicted tissue boundary shown in red and the actual tissue boundary in yellow.}
\vspace{-0.5cm}
\label{Exp_A}
\end{figure*}

To obtain the optimal $q_{x,0}^*$, the inequality constraints are first implicitly incorporated into the objective function:
\begin{align}
    I_-(q_{x,0}, \mathbf{q}) = 
    \begin{cases}
    +\infty, & F^\Delta \geq 0, \\
    -(1/t)\log(-F^\Delta), & F^\Delta < 0,
    \end{cases}
\end{align}
where $ F^\Delta = \hat{F}^{\max}(q_{x,0}, \mathbf{q}) - F^\mathrm{safe} $, and the approximate optimization function is:
\begin{align}
    \underset{q_{x,0}, \mathbf{q}}{\mathrm{minimize}} &\left(\hat{\mathbf{w}}^v(q_{x,0}, \mathbf{q}) - {}^E\mathbf{w}^b_0 \right)^T \left(\epsilon \frac{\partial \mathcal{L}_\mathcal{D}}{\partial \mathbf{w}^v} + (1-\epsilon) \frac{\partial \mathcal{L}_\Omega}{\partial \mathbf{w}^v} \right) \nonumber \\
    &+ I_-(q_{x,0}, \mathbf{q}).
\end{align}
Then, the genetic algorithm is employed to find the optimal solution $q_{x,0}^*$ for this optimization problem. The grasping position is selected during the offline phase, and it needs to be recalculated for different boundaries and ROIs. After the grasping position is determined, the grasper is bound to the tissue, and subsequent actions are decided by the controller.

\vspace{+0.25cm}
The complete process of the learning-based adaptive control method for exposure tasks is summarized in Algorithm \ref{alg}.
\vspace{-0.25cm}

\begin{algorithm}[!h]
    \caption{Learning-Based Adaptive Control}
    \label{alg}
    \renewcommand{\algorithmicrequire}{\textbf{Input:}}
    \renewcommand{\algorithmicensure}{\textbf{Output:}}
    \begin{algorithmic}[1]
        \REQUIRE Visual boundary parameters $\mathbf{w}^v_t$
        \ENSURE Control policy (robot's next state) $\mathbf{q}_{t+1}$
        
        \STATE Train deformation estimation model
        \STATE Select initial grasping pose $\mathbf{q}_0$ via genetic algorithm
        \STATE Sample around {$\mathbf{q}_0$} to obtain {$\mathbf{Q}_0$} 
        \STATE Estimate deformation {${}^E \hat{\mathbf{W}}^b_0$} 
        \STATE Compute initial Jacobian {$\hat{\mathbf{J}}_0 = {}^E\hat{\mathbf{W}}_0^b \mathbf{Q}_0^T \left( \mathbf{Q}_0 \mathbf{Q}_0^T \right)^{-1}$}
        \WHILE {$\mathcal{D} \neq \varnothing$ (until exposure)}
            \STATE Observe current visual boundary $\mathbf{w}^v_t$
            \STATE Estimate error {$ \mathbf{e} = \hat{\mathbf{J}}_{t-1} \delta \mathbf{q}_{t-1} - \delta \mathbf{w}_{t-1}^v $}
            \STATE Update Jacobian {$ \hat{\mathbf{J}}_t = \hat{\mathbf{J}}_{t-1} - \lambda \mathbf{e} \delta \mathbf{q}^T_{t-1} $}
            \STATE Compute velocity {$ \mathbf{u}_t = \frac{d}{dt} \mathbf{q} = -\kappa_1 \frac{\partial\mathcal{L}}{\partial\mathbf{q}} - \kappa_2 \frac{\partial\mathcal{L^F}}{\partial\mathbf{q}} $}
            \STATE Limit velocity {$ \mathbf{u}^{*}_t = \min \left( \mathbf{u}_t, s \cdot \mathbf{u}_t / \| \mathbf{u}_t \|_2 \right) $}
            \STATE Update controller {$ \mathbf{q}_{t+1} = \mathbf{q}_t + \mathbf{u}^{*}_t $} 
            \STATE Command grasper to reach {$\mathbf{q}_{t+1}$}
        \ENDWHILE
    \end{algorithmic}
\end{algorithm}

\section{Experiments}

\subsection{Experimental Setup}

To validate the performance of the proposed method, three experimental scenarios were designed. First, simulations were conducted in a virtual environment, where tissue deformation parameters remained identical to those used during training data acquisition for the deformation estimation network, while boundary geometries and ROIs were randomized. To align with real-world dimensions, the dimension of the simulated tissue was scaled to 20 cm.

Subsequently, real-world experiments employed two deformable specimens: polyvinyl alcohol (PVA) sponge \cite{karimi2014mechanical} and ex vivo porcine liver tissue. Compared to simulation, these materials exhibit distinct deformation characteristics, with the porcine liver further demonstrating complex geometry and tissue heterogeneity \cite{plantefeve2016patient}. All real-world experiments utilized the MP550 surgical system from United Imaging Surgical. One robotic arm mounted an endoscopic camera, prepositioned manually to ensure boundary visibility. The other arm was equipped with a grasper to clamp tissue boundaries and execute retraction under the control of the proposed algorithm.

To capture ground-truth deformation data, an Orbbec Gemini 336 stereo camera and an ATI Nano25 SI-250-6 force sensor were deployed. The camera reconstructed the surface point cloud and identified boundary points through color thresholding segmentation for real-time boundary estimation. The force sensor was installed beneath the support platform and was used to measure the force applied to the tissue. The retraction force $F^r$ was derived from the principle of force equilibrium: $F^r \approx{\|\mathbf{F}^{\text{sensor}}\|_2} $, where $\mathbf{F}^{\text{sensor}} \in \mathbb{R}^3$ represents the three-dimensional force measured by the sensor. The real-world experimental setup is illustrated in Fig. \ref{exp_setup}.

\subsection{Deformation Estimation }

\begin{figure*}[!t]
\centering
\includegraphics[width=7.1in]{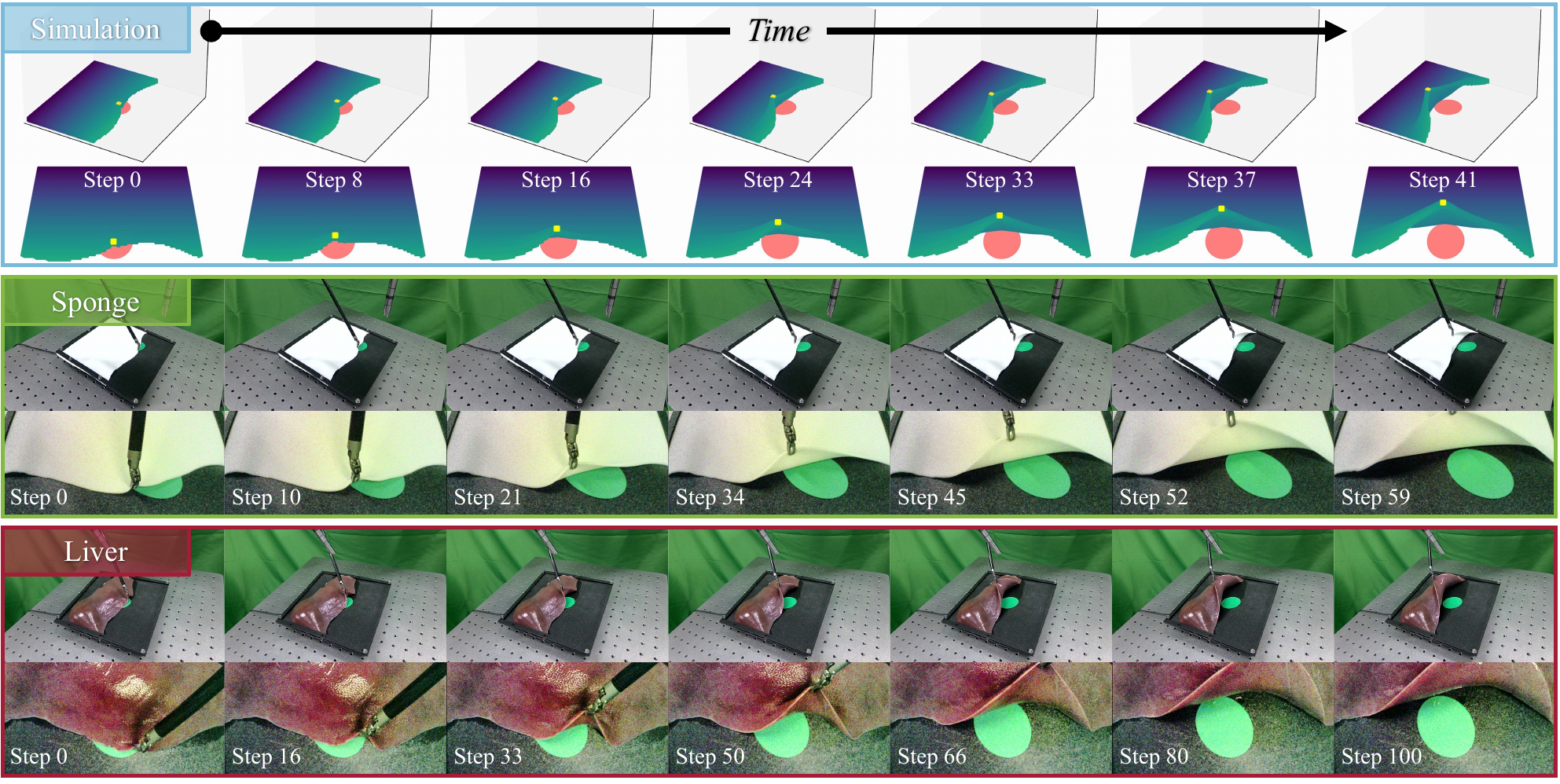}
\vspace{-0.7cm}
\caption{Snapshots from a representative autonomous retraction trial on simulation, PVA sponge, and porcine liver. Each set shows two views: the upper row displays the global perspective captured by an external camera, while the lower row presents the laparoscopic perception view.}
\vspace{-0.4cm}
\label{Exp_B}
\end{figure*}

\begin{figure}[!h]
\centering
\vspace{+0.05cm}
\includegraphics[width=3.45in]{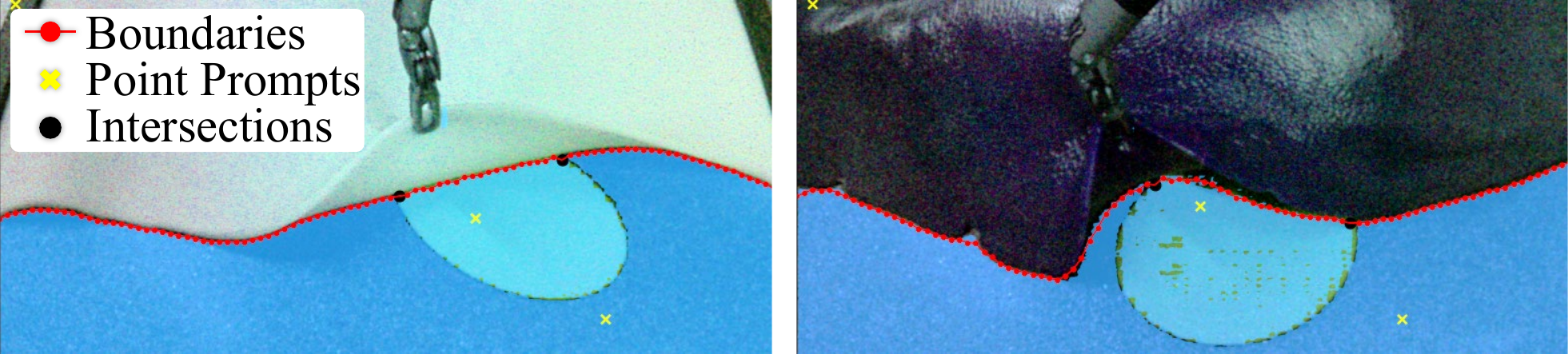}
\vspace{-0.7cm}
\caption{SAM2 was chosen to obtain the visual boundaries of the tissue.}
\vspace{-0.6cm}
\label{sam2}
\end{figure}

Before performing autonomous tissue manipulation, a deformation estimation network was trained and experimentally validated. The training data were collected as follows: the simulated tissue boundary was randomized 100 times; for each randomized shape, 10 grasping points were uniformly sampled along the boundary. For each shape-grasping-point pair, a 10-cm pulling action was applied in a randomly sampled direction. To ensure physical plausibility, the sampling range of the direction was constrained to the collision-free workspace. During each retraction, 100 time steps of data were recorded. This procedure yielded $10^5$ structured samples, each comprising the corresponding tissue boundary parameters, maximum deformation force, grasping location, and action.

The deep deformation estimation network was trained on the aforementioned simulated dataset. The loss function was defined as the mean squared error (MSE) between the predicted and ground-truth tissue deformation values. The MLP architecture employed LeakyReLU activation functions in all hidden layers and a linear activation in the output layer. Training was performed using the Adam optimizer \cite{kingma2014adam} with a batch size of 32, an initial learning rate of $1 \times 10^{-3}$, and default momentum parameters ($\beta_1 = 0.9$, $\beta_2 = 0.999$). The learning rate was dynamically halved via a ReduceLROnPlateau scheduler when the validation loss plateaued. Convergence was achieved after approximately 400 epochs. The dataset was partitioned into training and validation sets in a 9:1 ratio, and additional online evaluation was conducted within the simulation environment to assess real-time prediction performance.

To validate the performance of the deep deformation estimation network, a unified operational protocol was executed across three experimental scenarios: for tissues with varying geometric morphologies, retraction with a total displacement of 10 cm was applied from multiple grasping positions along a random direction. In each trial, the ground-truth tissue boundary was reconstructed from mesh or point cloud data and compared against the predicted boundary, which was generated by reconstructing the network’s output parameters. The mean closest distance (MCD) was adopted as the metric for prediction error, defined as:
\begin{equation}
    \text{MCD} (\mathcal{P}^b, \hat{\mathcal{P}}^b) = \frac{1}{N_{\mathbf{p}^b}} \sum_{\mathbf{p}^b \in \mathcal{P}^b} \min_ {\hat{\mathbf{p}}^b \in \hat{\mathcal{P}}^b} \|\mathbf{p}^b - \hat{\mathbf{p}}^b \| _2,
    \label{MCD}
\end{equation}
where $\hat{\mathcal{P}}^b$ denotes the predicted boundary, and $N_{\mathbf{p}^b}$ is the number of sampled points on the actual boundary. Since the predicted boundary is continuous and can be sampled at far more points ($\hat{\mathbf{p}}^b$) than the measured points ($\mathbf{p} ^b$), only the unidirectional MCD error is computed. A total of 1,000 simulations and 100 real-world experiments were conducted. Fig. \ref{Exp_A} presents the error distribution across all experiments and visualizes a retraction process from each scenario.

The results demonstrate that the model achieves low prediction errors in simulated scenarios that align with the training data distribution. Although the error increases on real tissues (including sponge and liver), it remains within a clinically acceptable range ($<$ 1 cm for retractions up to 10 cm), and the model accurately captures the overall deformation trend. Moreover, the model exhibits lower error under small displacement conditions, effectively supporting the adaptive control method, which relies on local sampling and inference around the initial configuration. In summary, this study provides preliminary validation that the proposed deep deformation estimation network can reliably support critical components of both simulation and real-world tasks.

\subsection{Adaptive Retraction}

\begin{figure*}[!t]
\centering
\begin{minipage}{0.163\linewidth}
    \centering
    \vspace{-0.24cm}
    \includegraphics[height=3.52in]{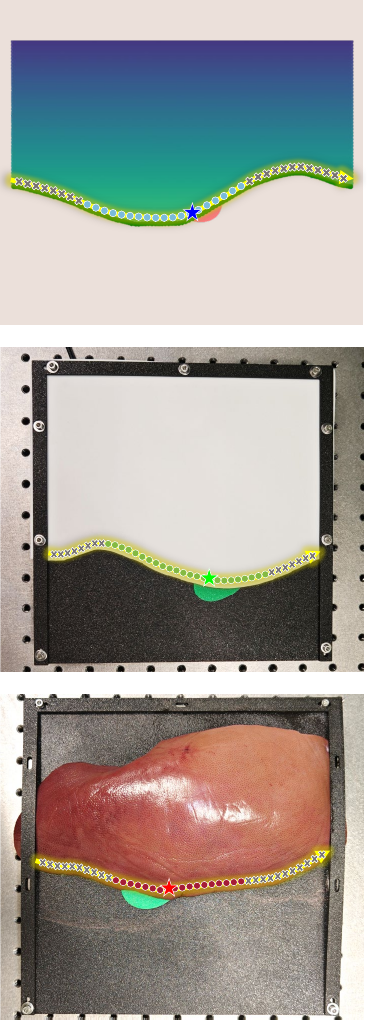}
\end{minipage}
\begin{minipage}{0.83\linewidth}
    \centering
    \includegraphics[height=3.66in]{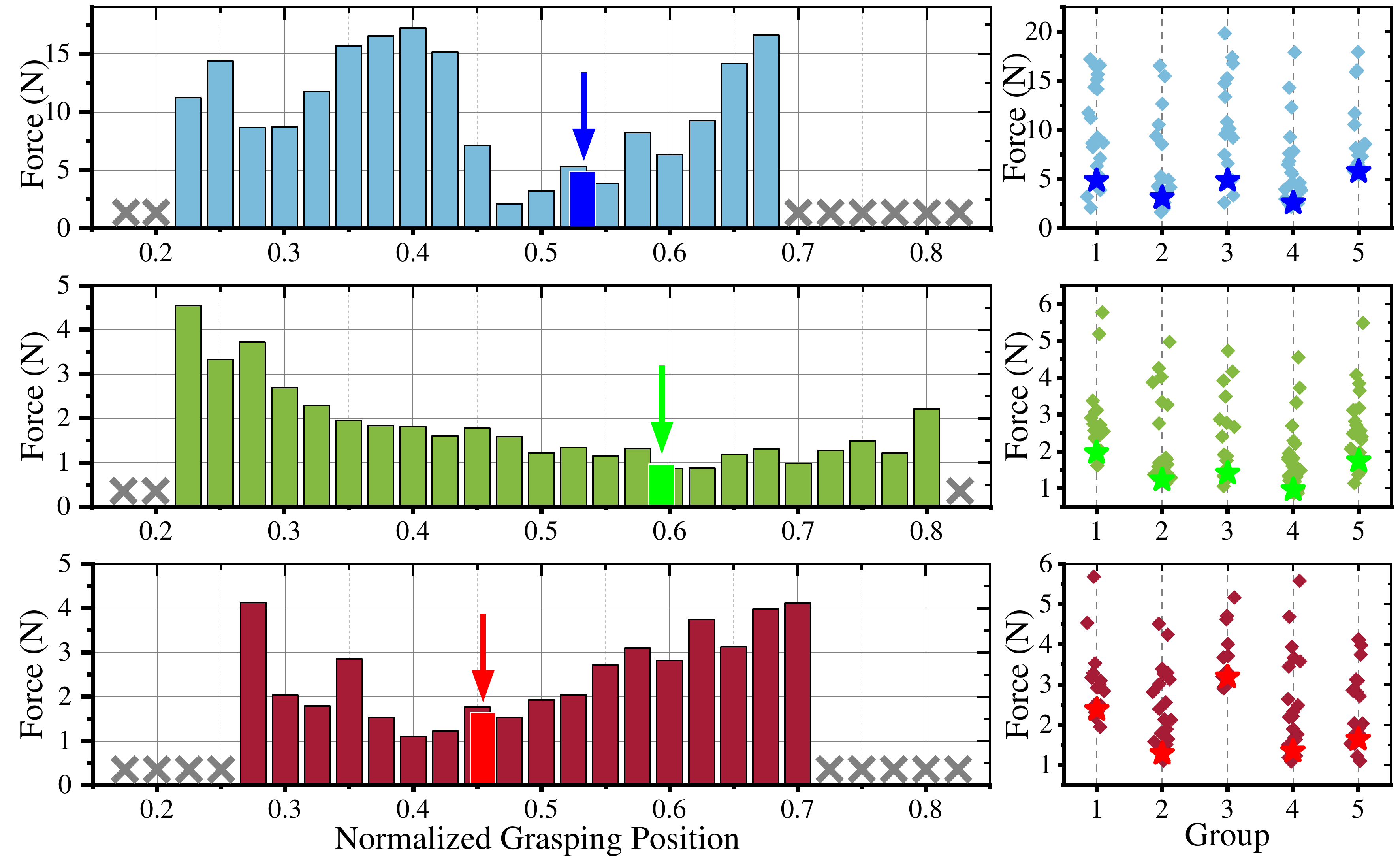}
\end{minipage}
\vspace{-0.4cm}
\caption{(Left) Spatial distribution of candidate grasping points along the tissue boundary. (Middle) Retraction forces required to achieve full ROI exposure when executing autonomous retraction from each candidate point (lower values are better). The grasping positions are normalized from left to right. The highlighted part corresponds to the grasping point selected by the proposed algorithm, while gray crosses indicate positions where exposure could not be completed under safety constraints. (Right) Statistical summary of retraction forces across five distinct configurations for each experimental scenario.}
\vspace{-0.52cm}
\label{Exp_C}
\end{figure*}

The proposed autonomous tissue retraction method was evaluated in three experimental scenarios. In simulation, the visual boundary ${}^E \mathcal{P} ^v$ serving as a state variable in the control loop is directly extracted from the mesh nodes of the tissue model. However, in real-world experiments, it must be obtained through an additional perception pipeline. To this end, we employ Segment Anything Model 2 (SAM2) \cite{ravi2024sam} to extract the visual boundary from laparoscopic images, as illustrated in Fig. \ref{sam2}. Specifically, SAM2 generates an image mask using a few offline-annotated point prompts, enabling the extraction of the tissue’s contour and the identification of its intersection points with the overlap area. Based on this boundary information, the optimal grasping position is computed offline via optimization and then manually set at the corresponding site prior to retraction. The controller then iteratively computes and executes motions to expose the ROI.

All experimental scenarios utilized the same deformation estimation network, trained exclusively on simulation data. However, since the closed-loop frequency and motion dynamics of the physical setup differed from those of the simulated experiments, we further adjusted the control parameters in the real-world experiments. These settings deliberately deviate from the simulation task configuration to enhance the algorithm’s performance, as detailed in Table \ref{param_table}.

\begin{table}[h]
    \vspace{-0.1cm}
    \caption{Hyperparameter configuration for experiments}
    \vspace{-0.4cm}
    \begin{center}
    \begin{tabular}{!{\vrule width1.2pt}c|c|c|c|c|c|c!{\vrule width1.2pt}}
    \Xhline{1.2pt}
    $ $ &$\kappa_1$ &$\kappa_2$ &$\lambda$ &$s^v$ &$s^\omega$ &$\epsilon$\\
    \hline
    Simulation   & 5e-3   & 1e-5   & 2e4    & 1e-3   & 1e-2   & 0.9 \\
    \hline
    Real-world   & 1e-2   & 1e-5   & 1e5    & 2e-3   & 1e-2   & 0.75 \\    
    \Xhline{1.2pt}
    \end{tabular}
    \end{center}
    \label{param_table}
    \vspace{-0.4cm}
\end{table}

Analysis of multiple retraction trials demonstrates that the proposed method completes the majority of tasks. In rare cases, such as when the ROI is placed in an extremely unfavorable configuration, the required deformation may exceed the maximum allowable deformation force constraint, potentially leading to exposure failure. Importantly, such failures arise from the physical infeasibility of the exposure task under those conditions, rather than from flaws in the method itself. For configurations deemed feasible a priori based on boundary geometry and ROI placement, the method achieves a success rate approaching $100\%$. Fig. \ref{Exp_B} presents snapshot sequences from a representative retraction trial in each experimental scenario, showing both a global view and a laparoscopic view. The latter provides an intuitive indication of the degree of ROI exposure achieved during the procedure.

\subsection{Grasping Point Selection}

To enhance retraction performance, the proposed method incorporates an optimized selection of grasping points. This set of experiments aims to evaluate the performance of the grasping point selection algorithm. Given a specific tissue boundary shape and region-of-interest configuration, the method computes an optimal grasping position using a genetic algorithm. For comparative purposes, we uniformly sampled candidate grasping points along the tissue boundary at 5 mm intervals over the 20 cm span, and included the genetically optimized point within this candidate set.

The full autonomous retraction procedure is then executed from each candidate grasping position, and the deformation force exerted on the tissue upon complete ROI exposure is recorded. Unlike in simulation, where internal nodal forces were directly accessible, such internal forces could not be measured directly in real-world experiments. Instead, a force sensor under the platform was used to indirectly measure the applied retraction force $F^r$. Since internal tissue deformation forces were generally positively correlated with the external loading, $F^r$ was adopted in this work as a surrogate metric for tissue deformation force.

For each experimental scenario, we conducted retraction trials under five different tissue and ROI configurations, recording $F^r$ each time the ROI was successfully exposed. Fig. \ref{Exp_C} presents the statistical distribution of retraction forces across these five configurations and provides a detailed breakdown of $F^r$ corresponding to all candidate grasping positions for one representative configuration.

The results show that, although the current grasping selection algorithm based on the deformation estimation network fails to precisely locate the globally optimal solution due to partial observability of the ROI and errors in network prediction results, the selected positions consistently lie within regions of relatively low retraction force. This demonstrates that, while not theoretically optimal, the proposed method effectively ensures operational feasibility and safety, making it a key enabler for stable and reliable autonomous retraction.

\section{Discussion and Conclusion}

This paper proposes a learning-based adaptive control framework for autonomous tissue retraction. The approach features a closed-loop adaptive controller that leverages the deformation Jacobian and is specifically designed for surgical exposure tasks. To improve autonomy and safety, a deformation estimation network is integrated to guide grasping point selection, determine initial configurations, and enforce safety constraints. Simulations and real-world experiments validate the effectiveness of each component, demonstrating that the proposed method achieves stable and reliable tissue retraction, showing potential for clinical applications.

As illustrated by the VPPV framework \cite{long2025surgical}, learning-based strategies can generate high-level trajectories, but precise execution still relies on servo controllers to execute low-level manipulation. Task-specific control methods remain a key factor in achieving reliable embodied intelligence, serving as a bridge between high-level general-purpose cognition and low-level physical execution. Building upon the successful implementation of tissue retraction, future work will explore unified modeling and control strategies for diverse surgical manipulations. Ultimately, these task-specialized modules will be integrated with high-level task planners and cognitive reasoning systems to enable safe, interpretable, and generalizable automation of surgical procedures.

\section{Acknowledgment}

We want to express our sincere appreciation to Wuhan United Imaging Surgical Co., Ltd. (UIS) for their generous help, not only giving us access to the MP550 robotic surgery system, but also providing constant technical support. We are particularly grateful to Dr. Xie Qiang, Dr. Yu Qiuyu, and Mr. Yang Fan for their expert guidance and hands-on assistance throughout this study. 

This work was supported in part by the Fundamental and Interdisciplinary Disciplines Breakthrough Plan of the Ministry of Education of China under Grant JYB2025XDXM208, in part by the Joint Funds of the National Natural Science Foundation of China under Grant U25A600011, and in part by Hubei Science and Technology Major Program under Grants 2023BCA002 and 2024BCB009.

\bibliographystyle{plainnat}
\bibliography{references}

\vspace{+5.5cm}
\centerline{\MakeUppercase{\appendixautorefname}}

\setcounter{section}{0}
\section{Parameterization of 3D Tissue Boundaries} \label{app: Parameterization 3D}

In the main text, we introduced the use of Wendland radial basis functions to parameterize the tissue boundary $\mathcal{P}^b$, which is compactly expressed as:
\begin{equation}
    \mathbf{W}^b = \mathcal{R}^b (\mathcal{P}^b).
\end{equation}
This appendix provides a detailed description of this parameterization process.

The general form of the radial basis function expansion is expressed as:
\begin{equation}
    \phi(\xi) = \sum_{j=1}^{m} w_j \, \varphi_{3,1} \left( \left|\frac{\xi - \xi_c^j}{h}\right| \right),
\end{equation}
where $w_j \in \mathbb{R}$ is the scalar weight, $\xi_c^j$ denotes the center of the $j$-th basis function, $h>0$ is the support radius, and $\varphi_{3,1}(\cdot)$ is the Wendland function yielding a $C^2$-continuous and compactly supported basis function. 

A key advantage of this basis function is that both its derivatives and integrals admit closed-form expressions, allowing for efficient and exact analytical computations in subsequent steps. Specifically, it is defined as:
\begin{equation}
    \varphi_{3,1}(|r|) = (1 - |r|)_+^4 (4|r| + 1), \text{with } r = \frac{\xi - \xi_c}{h},
\end{equation}
where $(\cdot)_+ = \max(0, \cdot)$. Its derivative with respect to $\xi$ is given by:
\begin{equation}
    \frac{d}{d\xi}\varphi_{3,1}(|r|) = -20 \, r (1 - |r|)^3_+ \cdot \frac{1}{h}.
\end{equation}
The definite integral of the basis function over an interval $[\xi_H, \xi_L]$ can also be evaluated analytically:
\begin{equation}
    \int^{\xi_H}_{\xi_L}\varphi_{3,1}(|r|) d\xi = \left(\text{sgn}(r_H)\mathcal{F}(|r_H|) - \text{sgn}(r_L)\mathcal{F}(|r_L|) \right) h,
\end{equation}
where $r_{H/L} = \text{clip}((\xi_{H/L} - \xi_c)/h, -1, 1)$, and $\text{sgn}(\cdot)$ returns +1, 0, or -1 depending on the sign of its argument. The antiderivative $\mathcal{F}(c)$ for $c \in [0,1]$ is:
\begin{align}
    \mathcal{F}(c) &= \int^c_0\varphi_{3,1}(|r|) dr \nonumber \\
                    &=(c-1)^5(4c+1)/5 - 2(c-1)^6/15 + \frac{1}{3}.
\end{align}
These closed-form expressions for the derivatives and integrals allow the control law to be formulated analytically, thereby avoiding numerical approximations.

\begin{figure}[!t]
\centering
\vspace{+0.1cm}
\includegraphics[width=3.48in]{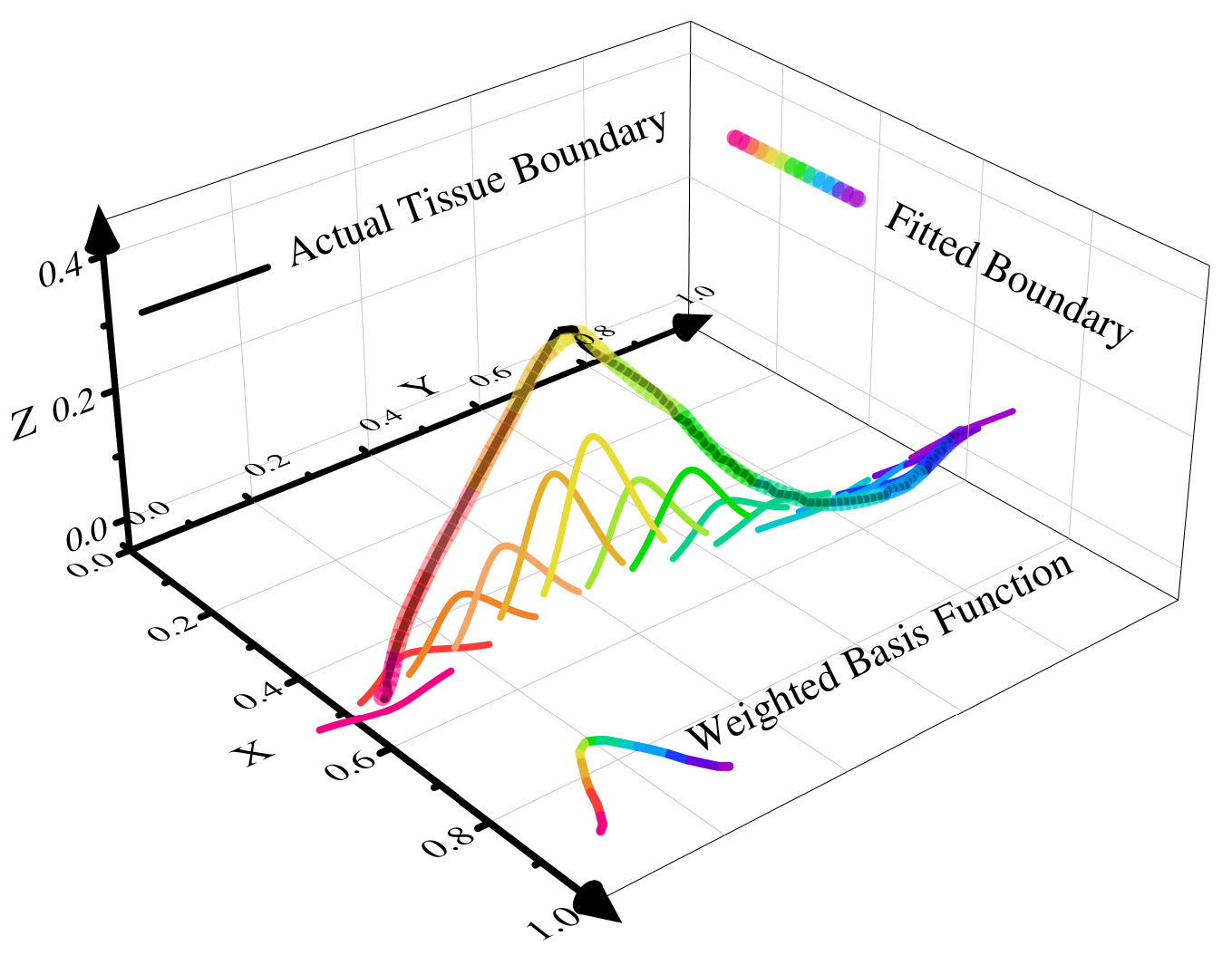}
\vspace{-0.6cm}
\caption{Visualization of 3D tissue boundary fitting using weighted cumulative basis functions. 
With $m = 15$, the fitted boundary (colored) closely matches the actual tissue boundary (black).}
\vspace{-0.2cm}
\label{3D_param}
\end{figure}

Then, the radial basis function is employed to parameterize $\mathcal{P}^b$. Data for $\mathcal{P}^b$ are collected during randomized simulation runs, comprising the coordinates of boundary grid nodes at each time step. These points are assumed to lie on the tissue boundary, denoted as:
\begin{align}
    \mathcal{P}^b_\mathrm{grid} = \{ \mathbf{p}^{b, i} = [x^{b, i},y^{b, i},z^{b, i}]^T \}^n_{i=1} \subset \mathcal{P}^b.
\end{align}
The parameterization proceeds by fitting these discrete points along each spatial dimension using $m$ basis functions of identical shape, uniformly distributed over the normalized parametric domain $\xi \in [0,1]$. Here, $\xi$ denotes the normalized parametric coordinate along the tissue boundary curve. To isolate deformation residuals and ensure coordinate consistency, the boundary points are centered with respect to a baseline curve defined by linear interpolation between the first and last boundary points:
\begin{align}
    \mathbf{p}(\xi) = (1-\xi)\mathbf{p}^{b,1} + \xi\mathbf{p}^{b,n}.
\end{align}
The residual displacement at each sample point is then given by:
\begin{align}
    \Delta \mathbf{p}^{b, i} = \mathbf{p}^{b, i} - \mathbf{p}(\xi^i),
\end{align}
where $\xi^i$ is the parametric coordinate corresponding to $\mathbf{p}^{b, i}$.
This leads to the following least-squares optimization problem:
\begin{equation}    
    \underset{\{ \mathbf{w}^j \}^m_{j=1}}{\mathrm{minimize}} \sum_{i=1}^n \left\| \Delta\mathbf{p}^{b,i} - \sum^m_{j=1}\mathbf{w}^j \varphi\left( \left | \frac{\xi^i - \xi_c^j}{h} \right | \right) \right\|^2_2 ,
    \label{minimize_1}
\end{equation}
where $\mathbf{w}^j \in \mathbb{R}^3$ is the weight vector associated with the $j\mathrm{th}$ basis function. Define the interpolation matrix $\mathbf{\Phi} \in \mathbb{R}^{n\times m}$ as:
\begin{equation}
    \mathbf{\Phi} = 
    \begin{bmatrix}
        \varphi\left( \left| \frac{\xi^1 - \xi_c^1}{h} \right| \right) & \cdots & \varphi\left( \left| \frac{\xi^1 - \xi_c^m}{h} \right| \right) \\
        \vdots & \ddots & \vdots \\
        \varphi\left( \left| \frac{\xi^n - \xi_c^1}{h} \right| \right) & \cdots & \varphi\left( \left| \frac{\xi^n - \xi_c^m}{h} \right| \right)
    \end{bmatrix},
\end{equation}
and let $ \mathbf{P}^b = [\Delta\mathbf{p}^{b,1} \cdots \Delta\mathbf{p}^{b,n}]^T $, $ \mathbf{W} = [\mathbf{w}^1 \cdots \mathbf{w}^m]^T $. Equation (\ref{minimize_1}) can then be expressed in matrix form as:
\begin{equation}
    \underset{\mathbf{W}}{\mathrm{minimize}} \left\| \mathbf{P}^b - \mathbf{\Phi}\mathbf{W} \right\| ^2_F.
\end{equation}
The least squares solution of the matrix equation is:
\begin{equation}
    \mathbf{W}^b = (\mathbf{\Phi}^T\mathbf{\Phi})^{-1}\mathbf{\Phi}^T\mathbf{P}^b ,
\end{equation}
where $\mathbf{W}^b \in \mathbb{R}^{m \times 3}$ constitutes the parameter matrix encoding the shape of the tissue boundary. Fig.~\ref{3D_param} shows the reconstructed boundary obtained by evaluating the weighted sum of basis functions using $\mathbf{W}^b$. The close agreement with the actual boundary validates the parameterization.

\section{Parameterization of 2D Visual Boundaries}

In the control iteration, in contrast to the parameterization used for $\mathcal{P}^b$, a distinct parameterization is employed for the visual boundary ${}^E\mathcal{P}^v$: 
\begin{equation}
    \mathbf{w}^v = \mathcal{R}^v ({}^E\mathcal{P}^v).
\end{equation}
This design choice arises from the fact that a 2D curve in the image plane admits a low-dimensional representation, while precise boundary coordinates are typically unavailable due to filtering artifacts in simulation or segmentation errors in real-world perception. Consequently, in the case of the simulation, a set of points near the evolving ${}^E\mathcal{P}^v$ is sampled instead:
\begin{equation}
    {}^E\mathcal{P}^v_\mathrm{near}=\{ {}^E\mathbf{p}^{v,k} = [{}^Ex^{v,k}, {}^Ey^{v,k}]^T \in \mathcal{V} ( {}^E\mathcal{P}_\mathrm{grid} ) \}^l_{k=1},
\end{equation}
where $\mathcal{V} ( {}^E\mathcal{P}_\mathrm{grid} )$ denotes the subset of grid points lying in the vicinity of the visual boundary ${}^E \mathcal{P}^v$. The curve parameters are then fitted to lie as close as possible to ${}^E\mathcal{P}^v_\mathrm{near}$, while ensuring that the reconstructed curve does not fall below any of them. This requirement is encoded as a set of inequality constraints, yielding the following constrained optimization problem:
\begin{align}
    & \mathbf{w}^v = \underset{{}^E\mathbf{w}}{\mathrm{minimize}}|| {}^E\mathbf{y}^v - \mathbf{\Phi}({}^E\mathbf{x}^v) {}^E\mathbf{w} || ^2_2. \\
    & \text{s.t.} \quad {}^k\left( \mathbf{\Phi}({}^E\mathbf{x}^v) {}^E\mathbf{w}\right) \geq {}^k\left({}^E\mathbf{y}^v\right), \quad \forall k \in \{1,\dots,l\},
\end{align}
where the corresponding coordinate vectors are ${}^E\mathbf{x}^v = [{}^Ex^{v,1} \cdots {}^Ex^{v,l}]^T$ and ${}^E\mathbf{y}^v = [{}^Ey^{v,1} \cdots {}^Ey^{v,l}]^T$. The interpolation matrix $\mathbf{\Phi}({}^E\mathbf{x}^v) \in \mathbb{R}^{l \times m}$ depends on the number of samples $l$, and ${}^k(\cdot)$ denotes the $k$th component of the vector. 

Due to the added inequality constraints, the optimization problem no longer admits a closed-form solution. Consequently, a dedicated convex optimization solver such as CVXPY is required. Additionally, this constrained formulation substantially increases the computational burden of the controller. In future work, it would be desirable to leverage more efficient and robust boundary-aware representations to alleviate this computational overhead. In any case, once this optimization problem is solved, the parameter vector $ \mathbf{w} ^v \in \mathbb{R}^m$ encoding the visual boundary ${}^E\mathcal{P}^v$ is obtained.

\end{document}